\documentclass{article}

    \usepackage[preprint,nonatbib]{neurips_2024}

\usepackage[utf8]{inputenc} 
\usepackage[T1]{fontenc}    
\usepackage{hyperref}       
\usepackage{url}            
\usepackage{booktabs}       
\usepackage{amsfonts}       
\usepackage{nicefrac}       
\usepackage{microtype}      
\usepackage{xcolor}         

\usepackage{tabularx}
\usepackage{color}
\usepackage{amsmath,amssymb,amstext}
\usepackage{multirow}
\usepackage{bbm}
\usepackage{bm}
\usepackage{wrapfig}

\usepackage{mathtools}

\usepackage{algorithm}
\usepackage{algorithmic}
\DeclareMathOperator*{\R}{\mathbbm{R}}
\usepackage{enumitem}

\usepackage{multirow}

\title{Leveraging Activations for Superpixel Explanations}

\author{%
	Ahcène Boubekki \\
	Physikalisch-Technische Bundesanstalt, Germany\\
	\texttt{ahcene.boubekki@ptb.de} \\
	\And
	Samuel Fadel \\
	Linköping University, Sweden\\
	\texttt{samuel.matthiesen@liu.se} \\
	\AND
	Sebastian Mair\\
	Uppsala University, Sweden \\
	\texttt{sebastian.mair@it.uu.se} \\
}

\begin{document}

	\maketitle

	\begin{abstract}
		Saliency methods have become standard in the explanation toolkit of deep neural networks. Recent developments specific to image classifiers have investigated region-based explanations with either new methods or by adapting well-established ones using ad-hoc superpixel algorithms. In this paper, we aim to avoid relying on these segmenters by extracting a segmentation from the activations of a deep neural network image classifier without fine-tuning the network. Our so-called Neuro-Activated Superpixels (NAS) can isolate the regions of interest in the input relevant to the model's prediction, which boosts high-threshold weakly supervised object localization performance. This property enables the semi-supervised semantic evaluation of saliency methods. The aggregation of NAS with existing saliency methods eases their interpretation and reveals the inconsistencies of the widely used area under the relevance curve metric.
	\end{abstract}

	\section{Introduction}
	
	The development of explainable AI (XAI) has accompanied the emergence of regulations regarding the use of machine learning models, especially for safety-critical applications such as healthcare~\cite{borys2023explainable} or surveillance~\cite{williford2020explainable,panfilova2024applying}. 
	Saliency methods hold an important role in the XAI toolkit owing to the problem they aim to solve: to unravel the influence of the input's features on the model's predictions.
	For image classifiers, the task is to highlight the most salient pixels, typically visualized as heatmaps.
	However, one limitation is that these methods may treat adjacent pixels independently, leading to discontinuous heatmaps that might make them sensitive to adversarial attacks~\cite{dombrowski2019explanations}.
	Masking-based methods do not have this weakness but at the cost of a lower resolution~\cite{RISE}.
	A compromise is to average the pixel methods over superpixels~\cite{XRAI,SWAG}. 
	{This so-called \emph{superpixelifcation} comes with the question of the selection of the superpixel alogrithm, and in particular, of their capability to produce a semantically relevant segmentation. }
	In this paper, we address both questions with a novel superpixel method that leverages the activation features of the image classifier to explain.
	
	{A deep neural network image classifiers are able to disentangle the semantics of an image, such as scene layout or object boundaries.}
	This has been studied in recent works of Caron et al.~\cite{DINO} on visual image transformers (ViT)~\cite{ViT} and of Kauffmann et al.~\cite{jacob} on convolutional networks (convets). 
	On convnets, different resolutions of the image semantics can be achieved depending on the depth from which feature activations are extracted.
	This property is the foundation of pyramid networks developed for tasks such as supervised and unsupervised object detection and semantic segmentation~\cite{FPN,UNET,FCOS}.
	Building on this concept, we demonstrate that clustering the feature activations of a convnet can yield a superpixel segmentation of the input image that is semantically aware.
	Note that the generation of superpixels needs to be unsupervised as our goal is to explain an image classifier without any form of fine-tuning, let alone fine-tuning for segmentation.
	
\begin{figure*}[!t]
		\centering
		\includegraphics[width=\textwidth]{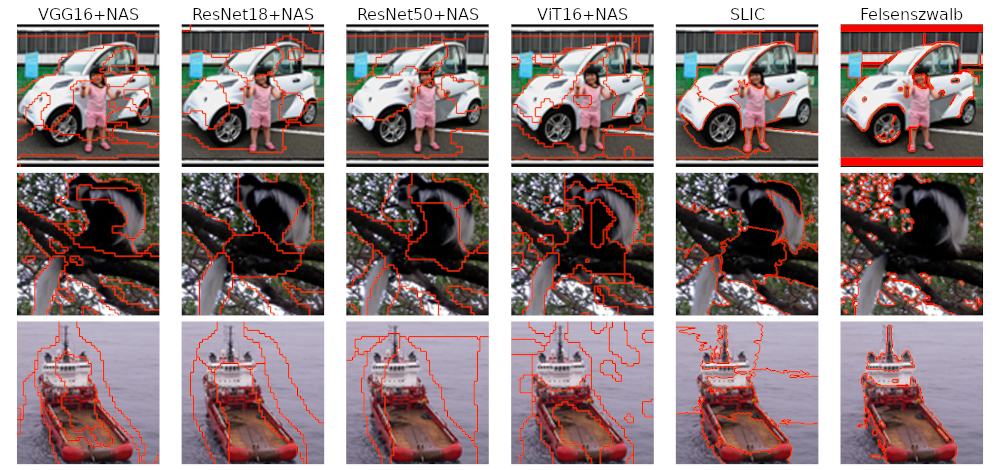}\\
		\caption{Examples of segmentations of STL-10 images adjusted to return a similar number of superpixels.
			The proposed NAS algorithm does not miss the tail of the monkey (second row) and does not artificially split the sea around the ship (third row). 
			If the cells including the person (first row) follow her shape, they also include both wheels. 
			Model agnostic SLIC and Felzenszwalb algorithms are fooled by the multiple edges.}
		\label{fig:intro}
	\end{figure*}

	{	Our concern about ad-hoc superpixel algorithms, is that they may overlook the semantics relevant to the model's prediction and thus can misguide the explanations. }
	Figure~\ref{fig:intro} illustrates this phenomenon for several image classifiers trained on STL-10. 
	{The segmentations based on our method (NAS) do not assign a superpixel to the person.}
	This is actually comforting since STL-10 does not have a class for humans.
	It means that the model does not \emph{see} the {person} as a piece of relevant information that helps to classify the image correctly. 
	Ad-hoc methods, such as SLIC~\cite{SLIC} and Felzenszwalb~\cite{FZ}, respectively used in SWAG and XRAI, follow the edges and isolate parts of the person and other irrelevant parts of the image, which might mislead the explanation of the predictions.

	{We propose, in this paper, to segment an image by clustering an image classifier's feature activations.}
	We call this method Neuro-Activated Superpixels (NAS) (Section~\ref{sec:method}).
	{	A class-wise clustering of the feature activations confirms that the resulting segmentation is semantically aware of the characteristics of each class. Building upon this result, we introduce a semi-supervised quantitative evaluation of saliency methods  (Section~\ref{sec:exp-sem}).}
	We then combine NAS superpixels with saliency methods and show how the resulting quantization eases the interpretation of saliency maps (Section~\ref{sec:exp-spx}).
	A qualitative and quantitative analysis reveals inconsistencies in the area under the least relevant curve metric (Section~\ref{sec:auc}). Finally, we show how the weakly supervised localization task indirectly evaluates the capabilities of NAS at extracting relevant semantics  (Section~\ref{sec:wsol}).

	\section{Related Works}
	We discuss three topics related to the problem and our approach. First, we position this work in the current development of saliency methods. Then, we discuss the computation and usage of superpixels in computer vision. Finally, we give a brief overview of the use of feature activations for image segmentation. Although this problem is beyond our scope, for the sake of completeness, we give a brief overview of how this technique is used {for image segmentation.}

	\subsection{Saliency Maps}
	Saliency methods can be split into two families. The first consists of model-agnostic methods {that estimate the sensitivity of the predictions to the perturbations of certain features of the input~\cite{zeiler2014visualizing,RISE,RELAX,LIME}.}
	{For example, RISE~\cite{RISE} generates saliency maps that are averages of a large number of occluding masks weighted with respect to how much they worsen the predictions. }
	The method is robust to pixel-based adversarial attacks, but requires the evaluation of a larger number of masks to produce a low-resolution saliency map.
	A different approach is represented by LIME~\cite{LIME}, which aims to decipher the relationships between the input and output with a linear approximation. Despite the theoretical guarantees applying only to simple architectures, LIME performs well on complex models. The theoretical background relies on the notion of neighborhood in the image space, which is still not well-defined and is not invariant to the dataset and task at hand.
	
	The second approach to saliency maps consists of model-aware methods that utilize the model's parameters {to compute the explanations}. These methods typically involve backpropagating the prediction to the input~\cite{GP,IG,GC,LRP}{, which is usually fast since just a single pass is needed.} The resolution is now at the pixel level, which has the drawback of making these methods sensitive to adversarial attacks~\cite{etmann2019connection}. 
	A workaround is to involve some perturbation of the input~\cite{smoothgrad} at the cost of increasing the computation time.
	Another solution, which is the one we employ here, is to average the saliency maps over superpixels~\cite{XRAI,SWAG}. The idea builds upon RISE's: compute several model-agnostic segmentations of the input, average the {pixels importance} computed using a backpropagation method over the superpixels, and return a weighted average. 
	Two representatives of this approach are XRAI~\cite{XRAI} and SWAG~\cite{SWAG}, which rely, respectively, on the Felzenszwalb algorithm~\cite{FZ} and a modification of Simple Linear Iterative Clustering (SLIC)~\cite{SLIC} for segmenting the input.
	{As discussed in the introduction, our concern is on capacity of these superpixel algorithms to capture semantics aligned with what the model learns. }
	
	\subsection{Superpixels}
	
	Superpixels are connected components of a partition of an image's pixels. 
	Ongoing development aims to go beyond color compression, as used in the GIF, to include more information, such as spatial proximity.
	
	Representative superpixel methods include Watersheds~\cite{Watersheds} and QuickShift~\cite{QuickShift}, which rely on specific clustering objectives. 
	The Simple Linear Iterative Clustering (SLIC)~\cite{SLIC} instead uses a spatially constrained $k$-means, which allows the control of the size of the superpixels.
	Another line of methods represents the augmented pixels as graph nodes, restating the task as a graph cut problem. Two examples of that family are the Normalized Cut~\cite{NormalizedCuts} and the Felzenszwalb~\cite{FZ} algorithms. The latter is used in XRAI. One drawback of that latter method is the limited control of the user over the number and, thus, size of the superpixels.
	Our Neuro-Activated Superpixels method also relies on $k$-means clustering {but} of the feature activations. This means that the size and resolution of the superpixels are controlled by the number of clusters and the depth of the feature activations.

	\subsection{Feature Activations}
	
	Integrating the encoder's feature activation in the decoder was key to the success of the U-Net architecture~\cite{UNET} as a general-purpose image {segmenter}.
	A similar idea has been used for convnet object detectors~\cite{FPN,zhang2020object} and semantic segmentation~\cite{abdal2021labels4free,cho2021picie}.
	{Several recent works have shown that it is possible to extract input semantics from a deep neural network classifiers feature activations.}
	The seminal work of Caron et al.~\cite{DINO} shows that the self-attention of the class token of self-supervised ViT16 highlights part of the image deemed related by the network. For convolutional neural networks, several works have investigated the presence and extraction of semantics in the activations~\cite{MaxAct,DeepDream}, revealing that deeper convolutional layers act less as edge detectors and capture more higher-level concepts, like objects.
	
	{A corollary is }that deeper convolutional layers have larger receptor fields and produce higher-dimensional activations. 
	This means that a naive aggregation of the feature activations weighs too much {the} lower layers, not necessarily translating meaningfully at the image level. On the other hand, if the first layers' activations are too strongly weighted, the superpixels will closely follow the picture edges, overlooking the information captured by the deeper layers.

	\section{Neuro-Activated Superpixels}\label{sec:method}

	\begin{figure*}[!t]
		\centering
		\includegraphics[width=\textwidth]{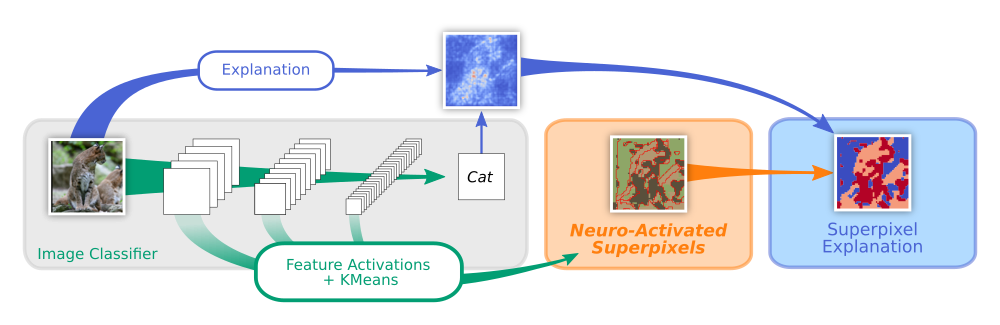}\\
		\caption{Schematic representation of our Neuro-Activatied Superpixels algorithm. The feature activations of a deep neural network image classifier are concatenated and then clustered, producing a segmentation of the input faithful to its semantics and internal operations of the network.}
		\label{fig:nas}
	\end{figure*}
	
	In this section, we introduce our Neuro-Activated Superpixels method, which returns superpixels aligned with the semantics captured by the model. Figure~\ref{fig:nas} depicts a schematic representation of our method.
	
	Let us consider a convolutional neural network $f$ trained for the classification of images of dimension $(H,W,C)$ into $Q$ classes, i.e., $f: \R^{H\times W \times C} \rightarrow \R^Q$. 
	The architecture of $f$ is usually a sequence of $L>0$ convolutional layers $\{l_i: \R^{h_i \times w_i \times c_i} \rightarrow \R^{h_{i+1}\times w_{i+1} \times c_{i+1}}\}_{1\leq i \leq L}$, followed by a multi-layer perceptron as a classifier, $p: \R^{h_{L+1}\times w_{L+1} \times c_{L+1}} \rightarrow \R^Q$. The operations of $f$ on an image $x \in  \R^{H\times W \times C}$ can thus be decomposed as follows:
		$f(x) = p \circ l_{L} \circ \ldots \circ l_1(x)$.
	In practice, $l_i$ can involve more than one convolution operation. For example, in the case of a ResNet, it can be a residual block.
	
	\begin{algorithm}[!h]
		\caption{NAS Python pseudocode}\label{alg:nas}
		\begin{algorithmic}[1]
			\STATE \tt x : input image of dimension (H,W,C)
			\STATE H,W: dimension of the output
			\STATE K: number of clusters
			\STATE{}
			\STATE  activations = []
			\FOR {\tt i = 1 to L}
			\STATE x = l${}_i$(x)
			\STATE u = Upsample(size=(H,W), mode=bicubic)(x)
			\STATE u = Reshape(u, (H*W,c${}_{i+1}$))
			\STATE u = u / Sqrt(Sum(u**2, axis=1))
			\STATE u = u / (1 + c${}_{i+1}$)
			\STATE activations.append(u)
			\ENDFOR
			\STATE activations = concatenate( activations, axis = 1)
			\STATE superpixels = Kmeans(K).fit-predict( activations )
			\RETURN superpixels
		\end{algorithmic}
	\end{algorithm}

	The operations of NAS are described in a Python-inspired pseudo-code in Algorithm~\ref{alg:nas}\footnote{A public implementation will be provided upon acceptance.}. For the sake of legibility, we extract all the feature activations here. However, as we shall see later, a subset of them may produce better results depending on the use case.
	Similarly, the choice of the output dimension $(H,W)$ compromises between a higher resolution and a {shorter computation time, most of it being dedicated to the clustering.}
	Note that in line 8, the upsampling uses a bicubic interpolation. However, if the final segmentation needs to be upsampled, we recommend using a nearest-neighbor interpolation. 
	{The clustering is computed using $k$-means for its speed and simplicity. Drawbacks include the fact that $k$-means assumes a uniform prior on the clusters' distributions, which may affect the relevance of the superpixels. Also, it returns inconsistent results that vary with the initialization. A workaround is to use hierarchical clustering, which is deterministic but heavy to compute.}
	
	\section{Evaluations}
	
	We start the evaluation by reviewing the different choices made in the definition of our algorithm and demonstrate that NAS extracts meaningful and relevant semantics {reflecting the classifiers internal choices and a hierarchy thereof.}.
	Next, we study the benefits of NAS "\emph{superpixelification}" of classical saliency methods both qualitatively and quantitatively. 
	Finally, we show that NAS superpixels improve these methods' Weakly Supervised Object Localization (WSOL)\cite{wsol} performance, especially at high overlapping thresholds.

	\subsection{Experimental Setting}\label{sec:exp-set}
	The focus of this set of experiments is to show that NAS captures relevant semantics and how it pairs with saliency methods. Hence, we involve a large spectrum of saliency methods but limit ourselves to two datasets.
	
	{\bf Baselines}
	We compare NAS superpixels with that of SLIC~\cite{SLIC} and the Felsenszwalb algorithm~\cite{FZ}, using their skimage implementations~\cite{skimage}.
	In terms of pixel-based saliency methods, we use Integrated Gradients (IGrad)~\cite{IG}, LRP~\cite{LRP}, GCAM++~\cite{GC}, and RISE~\cite{RISE}.
	The superpixels-based saliency baseline is XRAI~\cite{XRAI}, for which we rely on the code provided by the authors. 
	
	{\bf Architecture}
	We compare four common architectures: VGG16~\cite{VGG}, ResNet18, ResNet50~\cite{ResNet}, and ViT16~\cite{ViT}.
	All models rely on the standard torchvision's pre-trained implementations~\cite{pytorch}.
	We extract feature activations at five depth levels indexed from 0 to 4.
	For VGG16, we extract convolution activations preceding each max-pooling operation (without ReLU).
	For ResNet18 and ResNet50, we extract activations after the first max-pooling (input of the first residual block), followed by the output of the four residual blocks.
	For ViT16, we extract the activations of every second encoder starting from the third.
	
	{\bf Datasets}
	If not indicated otherwise, all the visual examples are drawn from the STL-10 dataset~\cite{coates2011analysis}. It is similar to ImageNet but smaller, facilitating the reproducibility of our experiments. It consists of 5000 images for training and 8000 images for testing of size 96$\times$96 split into ten classes.
	Similarly, the WSOL performance is evaluated solely on the CUBv2~\cite{WahCUB_200_2011} dataset, which consists of 5994 training images, 1000 validation images, and 5794 test images, all accompanied by a bounding box indicating the location of a bird belonging to one of the 200 classes. 
	
	{\bf Hardware}
	We ran all experiments on a single NVIDIA L40S GPU with 45GB memory which is in a machine that has a 32-core AMD 7452 CPU and 500 GB of RAM.	
	
	\subsection{Superpixels Evaluation}
	
	In the first set of experiments, we study the influence on the superpixels of the different parameters, namely, the backbone architecture, the depth of the feature activations, and the number of clusters, {revealing a certain hierarchy in the semantics preserved by the model.}
	We extend the analysis with an ablation study of the scaling and weighting operations (lines 10 and 11 of Algorithm~\ref{alg:nas}) in Section~\ref{apx-sec:abl} of the appendix.

	\subsubsection{Architecture}
	Superpixels produced by NAS on four standard computer vision models, VGG16, ResNet18, ResNet50, and ViT16, are shown in Figure~\ref{fig:intro}. 
	The segmentations are computed based on the last three extracted feature activations and clustered into five groups (K=5).
	For completeness, we add the superpixels produced by SLIC and Felzenszwalb and parametrize them to return a number of superpixels similar to NAS'.
	
\begin{figure*}[!t]
		\centering
		\includegraphics[width=.95\textwidth]{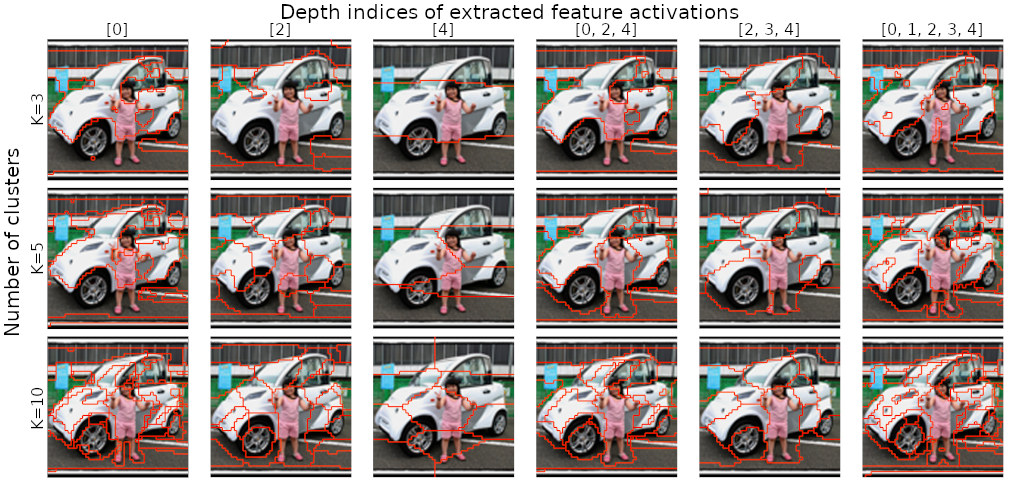}
		\caption{Ablation study of the influence of the extracted feature activations' depth and of the number of clusters. The shallower the activations, the more the superpixels fit to the image's strong edges.}
		\label{fig:KvsLayer}
	\end{figure*}

	Both model-agnostic methods, SLIC and Felzenszwalb, are too dependent on the images' edges. This applies especially to Felzenszwalb, which dedicates many superpixels to the background and details with little relevance, such as the car's door handle (first row). For the ship (third row), the internal rules of the spatial clustering of SLIC cause the grid-like partition of the water. 
	On the other hand, NAS superpixels focus on higher-level concepts. For example, the monkey's white tail always has its own superpixel (second row), which shares the same cluster with the white fur of the monkey's back. If the superpixels covering the person (first row) do follow her shape, they include her with the wheels, which are more relevant for the correct prediction. In the next section (Figure~\ref{fig:KvsLayer}), we show that if we reduce the number of clusters, the person is forgotten as the method splits foreground and background. If we increase the number of clusters, the person is in her own superpixel.  
	
	We see in Figure~\ref{fig:intro} that compared to convnets, ViT16 generates a lot of artifacts, which are especially visible in the picture of the ship. The stronger compression of ResNet50 yields semantically relevant superpixels yet more complex. For example, it splits the monkey into six cells, while VGG16 does it in three: two for the white fur and one for the dark fur. Yet, VGG16 is not as sensitive as ViT16, whose superpixels are the closest to the image's salient edges. Overall, ResNet18 returns a good trade-off between concept and edge detection. Hence, if not indicated otherwise, we continue the analysis using this architecture.

	\subsubsection{Depth and Number of Clusters}

	The resolution of NAS superpixels depends on the depth of the feature activations and the number of clusters. In Figure~\ref{fig:KvsLayer}, we fix the architecture to a ResNet18 and consider different combinations of a number of clusters (K) and feature depths. The latter are indexed from 0 (shallow) to 4 (deep). See Section~\ref{sec:exp-set} for more details.
	
	Using only the first depth ([0], first column), the borders of the cells closely follow the edges of the image. 
	As the number of clusters increases, the partition refines; however, the clustering remains focused on color and luminance changes.
	As the depth of the layers increases, the clusterings simplify but also become more abstract.
	Combining different depths compromises edge and concept detection. For example, NAS based on layers [0,2,4] or [0,1,2,3,4] isolates the person for K=5. With layers [2,3,4], it happens for K=10. This difference suggests that the concept represented by the ``\emph{human}" is not that salient in the deeper layers, meaning that it is not that relevant for the model to achieve its task: classifying STL-10, which does not have a ``\emph{human}" class. {In other words, the model does learn a certain hierarchy of the input's image semantics in which the person is ranked lower than the wheels in order to assign the image the correct ``\emph{car}'' label.}
	
	Figure~\ref{fig:intro} is computed using the combination [2,3,4] and K=5 and shows a different partition of the car picture. This variation is due to the stochasticity of $k$-means, which is discussed in Sections~\ref{apx-sec:cons} and \ref{apx-sec:agg} of the Appendix. If not indicated otherwise, we use the combination of ResNet18, feature activation [2,3,4], and K=5 in the remaining, as it achieves the best trade-off to illustrate our method's capabilities.

	\subsubsection{Superpixels and Semantics}\label{sec:exp-sem}
	
\begin{figure*}[!h]
		\centering
		\includegraphics[width=\textwidth]{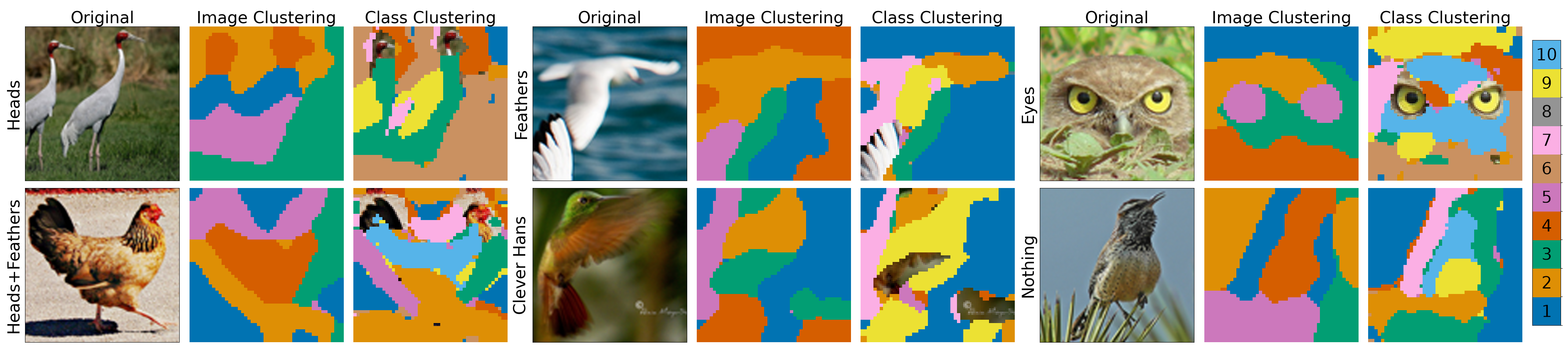}\\
		\caption{Clusterings learned on a single image or on all images of the same class are consistent with each other. Class-cluster~8 captures visually different but specific features of the birds, enabling a semi-supervised evaluation of the saliency methods.}
		\label{fig:semtc}
	\end{figure*}

	To show that our method can capture relevant class information, we extract feature activations of all the training images of the bird class of STL-10 and apply hierarchical clustering with Ward linkage, from which we extract 10 classes. Since this method cannot assign clusters to new data, we train a $k$-nearest neighbor (knn) classifier to learn the same partition. 
	In Figure~\ref{fig:semtc}, we show for six images the image-wise clustering using $k$-means with K=5 and the class-wise clustering computed from the aforementioned knn. For the latter, we uncover the superpixels of Cluster~8. Note that all the images are correctly assigned to the bird class by the underlying ResNet18 classifier.

	\begin{wraptable}[18]{r}{5cm}
		\vspace{-1em}
		\caption{Average saliency attributed to Class-Cluster~8 depending on the correctness of the predictions.}\label{tab:semtc}
		{	\small \centering
			\begin{tabular}{@{\extracolsep{\fill}}p{1.4cm}ccc}
				\toprule
				& Most				& 					& Other			\\
				Method 		& Salient	 		& Cluster~8 		& Clusters 	    \\  \midrule
				\multicolumn{4}{c}{Correctly Classified} \\ \midrule
				IGrad. & 4 & ${ 0.446  }$ & ${ 0.441  }$ \\
				LRP & 8 & ${ \bf 0.328  }$ & ${ 0.182  }$ \\
				GCAM++ & 10 & ${ 0.367  }$ & ${ \bf 0.428  }$ \\
				RISE & 8 & ${ \bf 0.535  }$ & ${ 0.458  }$ \\
				XRAI & 8 & ${ \bf 0.544  }$ & ${ 0.358  }$ \\ \midrule
				\multicolumn{4}{c}{Wrongly Classified} \\ \midrule
				IGrad. & 1 & ${ 0.418  }$ & ${ 0.413  }$ \\
				LRP & 8 & ${ \bf 0.312  }$ & ${ 0.195  }$ \\
				GCAM++ & 10 & ${ 0.375  }$ & ${ \bf 0.436  }$ \\
				RISE & 4 & ${ 0.494  }$ & ${ 0.468  }$ \\
				XRAI & 8 & ${ \bf 0.533  }$ & ${ 0.377  }$ \\
				\bottomrule	
		\end{tabular}}
	\end{wraptable}
	First of all, it is remarkable how similar the image-wise and class-wise clusterings are. {For all the selected examples, the} unmasked Cluster~8 of the class clusterings always has a counterpart in the image clusterings. 
	Since the class clusterings assigned the same labels to similar concepts captured by the network in different images, the analysis thereof is easier. 
	The background is either assigned Cluster~1 (dark blue) or 2 (orange). 
	Cluster~3 (green) seems to focus on vertical edges. {We uncovered Cluster~8 as} it turns out that it captures specific parts of the birds: head, tail, long feathers, eyes, but also Clever Hans~\cite{CleverHans} like the watermark (second row, second column). 
	Although this cluster is not present in all bird images, the fact that visually and semantically different parts of the birds are clustered together suggests that the model interprets them similarly as high-level concepts related to the bird class.
	This result opens the door to a semi-supervised quantitative evaluation of saliency methods. Indeed, once semantical clusters are identified on a subset of images, one can quantify how much saliency these methods attribute to these specific clusters.
	
	We report on such an evaluation in Table~\ref{tab:semtc}. 
	The saliency heatmaps for each test image are quantized using NAS superpixels based on the class clustering. 
	The average {saliency} of a class clustering is computed from that of all the superixels generated from that cluster.
	The cluster with the largest {saliency} is reported in the second column, the average {saliency} of Cluster~8 in the third, and the average {saliency} of all the other clusters in the last column.
	We split the test images depending on whether the class was correctly predicted in order to gain some insight into the relevance of Cluster~8 for the model's prediction.
	
	Two methods do not assign Cluster~8 as the most salient cluster among correctly classified images, and three methods for wrongly classified images. RISE is the only method that assigns more {saliency} to Cluster~8 when the prediction is correct and to another one when it is not.
	This divergence in behaviors highlights that saliency methods do not summarize the same information, which is also noticeable on the superpixel saliency maps.

	\subsection{Superpixel Explanations}
	
	We now study qualitatively and quantitively the \emph{superpixelification} of saliency maps.
	The final experiment demonstrates how combining saliency methods with NAS affects their WSOL performance, indirectly evaluating the capabilities of NAS to capture relevant semantics.

	\subsubsection{Superpixelification}\label{sec:exp-spx}
	
\begin{figure*}[!h]
		\centering
		\includegraphics[width=.95\textwidth]{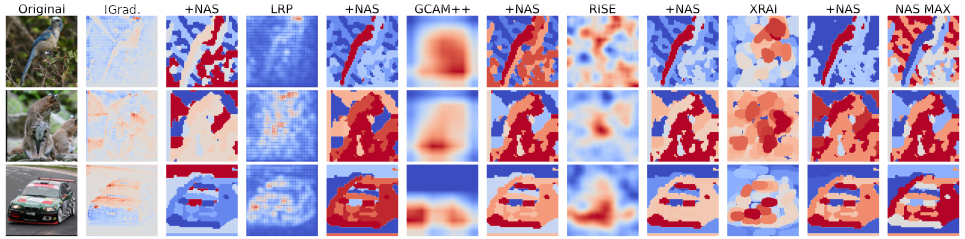}\\
		\caption{The superpixelification of the saliency maps contrasts the main object with the background, enabling analysis without looking at the original image.}
		\label{fig:expl}
	\end{figure*}

	The qualitative evaluation of saliency methods is usually done by superposing saliency heatmaps over the original image and then discussing the hot and cold spots.
	Superpixel heatmaps average the information per cell, facilitating the interpretation. We call the process \emph{superpixelification}.
	In Figure~\ref{fig:expl}, we show the NAS superpixelification of several saliency methods based on ResNet18, using feature activations 0 to 4 and K=5. To ease the comparison, heatmaps are normalized between 0 and 1.
	We also include a superpixel saliency map based on a greedy maximization of the Area Under the Least-Relevant First Curve (AUC-LeRF)~\cite{covert2021explaining,RISE}. Details of the maximization algorithm are given in the next Section~\ref{sec:auc}. {For this last approach, the} color gradient indicates the order in which the superpixels are hidden from first (blue) to last (red).

	The compromise between the scene layout and the model's regions of interest offered by NAS superpixels allows us to understand the scene without looking at the original image. In most cases, the main object of interest contrasts with the background, especially for GCAM++, RISE, and XRAI. Yet, the diversity represented here suggests that the method do not \emph{explain} the same internal mechanism of the model.
	
	It is interesting to see that for the bird image, all methods consider a part of the bird as the most salient, yet the maximization of the AUC masks the bird first.
	Following this process creates a negative image of the bird, which the model still recognizes correctly because of its shape. 
	Recall that we use a greedy maximization algorithm here and that for the two other images, the main object is deleted last. Nevertheless, this counterintuitive yet valid strategy casts doubt on the relevance of using the AUC-LeRF as an evaluation metric.

	\subsubsection{Quantification}\label{sec:auc}
	
	The area under the LeRF curve is nonetheless a standard evaluation metric of saliency methods~\cite{covert2021explaining,RISE}. 
	We report in Table~\ref{tab:lerf} average AUC-LeRFs in percentage over 100 test-images of each class {of STL-10, for a total of 1000 images}. We use the pre-softmax scores. We compare several saliency methods and their SLIC and NAS superpixelifications. The image's RGB channels are standardized before processing, and the masking operation sets the pixels to zeros.
	To allow a fair comparison, the curves are scaled to start and end on 1 and 0, respectively.
	For completeness, we include values obtained by greedy maximization of the AUC-LeRF using SLIC and NAS superpixels. Larger values are better. Best scores for the same saliency method and non-statistically different ones are marked in bold. 
	Figure~\ref{fig:lerf} depicts the average LeRF curves for each method.

	\begin{table*}[!h]
		\begin{minipage}{.45\linewidth}
			\caption{AUC-LeRF in percentage for saliency methods and their superpixelificaton.}
			\label{tab:lerf}
			{	\small \centering
				\begin{tabular}{@{\extracolsep{\fill}}p{1.3cm}cp{.1cm}p{1.2cm}c}
					\toprule
					
					{Methods} & ResNet18  & & {Methods} & ResNet18 \\ \midrule

					IGrad. & ${\bf 47.7 \pm 19.2 }$ 	& & RISE & ${\bf 64.1 \pm 47.2 }$ \\
					\hspace{1em}+SLIC & ${\bf 42.1 \pm 24.9 }$ 		& & \hspace{1em}+SLIC & ${\bf 71.1 \pm 25.1 }$ \\
					\hspace{1em}+NAS & ${\bf 46.9 \pm 31.1 }$ 		& & \hspace{1em}+NAS & ${\bf 76.9 \pm 53.4 }$ \\
					\midrule
					
					LRP & ${ 25.1 \pm 13.7 }$ 			& & XRAI & ${ 53.5 \pm 22.1 }$ \\
					\hspace{1em}+SLIC & ${\bf 69.3 \pm 19.1 }$ 		& & \hspace{1em}+SLIC & ${\bf 63.9 \pm 25.1 }$ \\
					\hspace{1em}+NAS & ${ 61.3 \pm 28.6 }$ 			& & \hspace{1em}+NAS & ${\bf 69.2 \pm 35.7 }$ \\
					\midrule
					
					GCAM++ & ${ 47.4 \pm 31.2 }$ 	& & \multicolumn{2}{c}{AUC Max.} \\ 
					\hspace{1em}+SLIC & ${\bf 65.9 \pm 49.6 }$ 		& & SLIC & ${\bf 87.8 \pm 41.0 }$ \\
					\hspace{1em}+NAS & ${\bf 65.9 \pm 40.5 }$ 		& & NAS  & ${\bf 91.1 \pm 52.6 }$ \\

					\bottomrule	
			\end{tabular}}
		\end{minipage}
		\hfill
		\begin{minipage}{.5\linewidth}
			\begin{figure}[H]
				\centering
				\includegraphics[width=\linewidth]{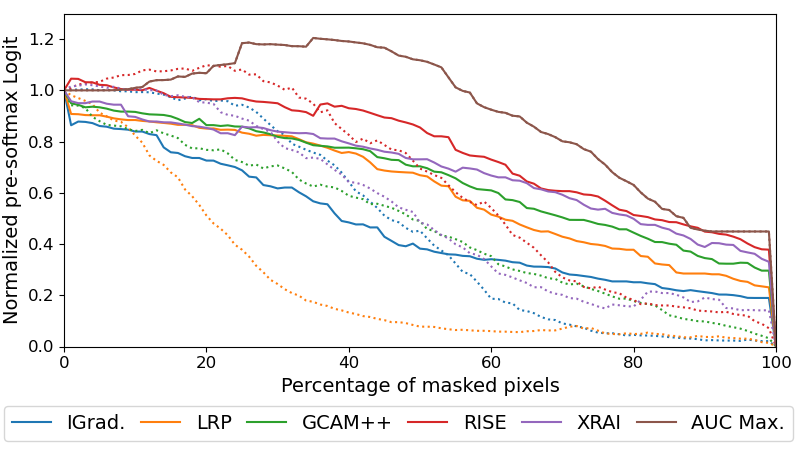}\\
				\caption{Average LeRF curves of saliency methods with (plain) and without NAS (dotted).}
				\label{fig:lerf}
			\end{figure}
		\end{minipage}
	\end{table*}

	Superpixelification using SLIC or NAS significantly increases the AUC for all the methods except Integrated Gradient (IGrad) and RISE. Yet, these scores remain at least 10 points smaller than those obtained by maximizing the AUC-LeRF based on SLIC or NAS superpixels.
	Although they produce different superpixels, SLIC and NAS return similar scores further challenging the relevance of the AUC-LeRF metric.

	\subsubsection{Weakly Supervised Object Localization }\label{sec:wsol}

	We build the last experiment based on the protocol exposed in \cite{wsol} and reuse their implementation and experimental setting for CUBv2\footnote{\url{https://github.com/clovaai/wsolevaluation}}.
	The metric used is {\tt MaxBoxAccV2} defined in \cite{wsol}, which is the average of the three bounding box Intersection over Union (IoU) at 30\%, 50\%, and 70\%.
	In Table~\ref{tab:wsol}, we report the four scores for a selection of saliency method as well the difference brought by SLIC and NAS test on ResNet18 using the last three feature activations and K=10. Results for other architectures can be found in Section~\ref{apx-sec:wsol}.
	
	\begin{wraptable}[21]{r}{5.5cm}
		\vspace{-1.5em}
		\centering
		\caption{Weakly Supervised Object \\Localization scores on CUBv2.}\label{tab:wsol}
		{	\small \centering
			\begin{tabular*}{\linewidth}{@{\extracolsep{\fill}}p{1.2cm}cccc}
				\toprule 
				&  {\tt MaxBox} & IoU 	& IoU 	& IoU\% \\ 
				Methods &  {\tt AccV2}  & @30\% & @50\% & 70\% \\ \midrule
				
				IGrad. & 36.7 & 68.8 & 31.4 & 9.9 \\
				\hspace{1em}+SLIC & +1.5 & +5.0 & +0.9 & -1.5 \\
				\hspace{1em}+NAS & +1.5 & -1.1 & +3.3 & +2.4 \\
				\midrule
				
				LRP & 63.6 & 95.4 & 71.3 & 24.0 \\
				\hspace{1em}+SLIC & -8.1 & -1.5 & -11.5 & -11.5 \\
				\hspace{1em}+NAS & +0.8 & -3.8 & -2.7 & +9.0 \\
				\midrule
				
				GCAM++ & 57.1 & 97.4 & 63.9 & 9.9 \\
				\hspace{1em}+SLIC & -1.1 & -1.0 & -3.4 & +1.1 \\
				\hspace{1em}+NAS & +8.9 & -3.0 & +8.6 & +21.1 \\
				\midrule
				
				RISE & 32.4 & 65.3 & 26.2 & 5.7 \\
				\hspace{1em}+SLIC & +0.7 & +0.9 & +1.2 & +0.1 \\
				\hspace{1em}+NAS & +6.4 & +2.6 & +9.4 & +7.0 \\
				\midrule
				
				XRAI & 31.3 & 63.2 & 24.9 & 6.0 \\
				\hspace{1em}+SLIC & +0.9 & +0.8 & +1.6 & +0.2 \\
				\hspace{1em}+NAS & +4.9 & +2.0 & +7.4 & +5.3 \\
				
				\bottomrule	
		\end{tabular*}}
	\end{wraptable}
	The NAS superpixelification improves the agglomerated score {\tt MaxBoxAccV2} for all five methods, whereas SLIC slightly improves only three.
	In detail, the effect of NAS is predominant on the most difficult metric, IoU@70\%, which counts how many predicted bounding boxes intersect 70\% of the ground truths.
	Every combination with NAS sees an increase of that score with a peak at +21.1 points with GCAM++. 
	On the other hand, NAS worsens, on average, the scores for the easier task of IoU@30\%, with a worse decrease of -3.8 points with LRP.
	Our interpretation is that the quantization brought by NAS superpixel improves the IoU@70\% because the method highlights the relevant part of the image. 
	The downside is that the heatmap is not smooth anymore, which {helps to improve} the less restrictive IoU@30\%.
	
	The intention behind this experiment is not to present NAS as a possible WSOL method but to show that this experiment can indirectly quantify how well the superpixelization of the saliency maps captures the relevant objects of the scene. The consistent improvement in the IoU@70\% scores brought by NAS for all saliency methods and architectures (See Section~\ref{apx-sec:wsol}) confirms the capabilities of our method to detect relevant semantics in the input. This point is reinforced by the fact that SLIC does not consistently improve that metric.

	\section{Conclusion}
	
	We have introduced Neuro-Activated Superpixels, a novel unsupervised image segmentation algorithm based on the clustering of feature activations of a deep neural network. The primary limitation of our approach is the use of $k$-means clustering, as it introduces some variability. Although hierarchical clustering could be a potential solution, it comes with increased computational cost. 
	
	The proposed method demonstrated its capability to segregate semantically meaningful regions within an image, which was confirmed by class-wise clustering. Furthermore, we presented a quantitative protocol for a semi-supervised evaluation of saliency methods based on the semantic significance they highlight. Among the methods evaluated, only RISE showed alignment between the activation of semantically relevant clusters and the model's predictions. Additionally, we demonstrated that the concept of superpixelification, or aggregating saliency maps by superpixel, simplifies the interpretation. When applied, this technique revealed inconsistencies in the AUC-LeRF metric, questioning its validity. Finally, the improvements that NAS brings to the WSOL performance of saliency methods indicate that it effectively captures semantically relevant features.

	\begin{ack}
This work was partially supported by the Wallenberg AI, Autonomous Systems and Software Program (WASP) funded by the Knut and Alice Wallenberg Foundation; as well as Sweden's Innovation Agency (Vinnova) project 2022-03023.
	\end{ack}

	\bibliographystyle{unsrt}
	\bibliography{ref}
	\newpage

	
	\appendix

	\section{Ablation Study}\label{apx-sec:abl}
	In the main text, we studied the influence of the depth of the feature activations and the number of clusters. We complete this analysis here with the ablation study of the scaling and weighting of the feature activations occurring in lines 10 and 11 of Algorithm~\ref{alg:nas}, respectively.
	
	ViT16 is not included here because the outputs of the several encoders it is made of return the same dimension and scaling. Hence, this model is not impacted.
	The scaling and weighting of the feature extension impact the other architectures more or less. For example, VGG16 yields good superpixels without these operations.
	Conversely, the quality and meaningfulness of the superpixels produced with ResNet18 and ResNet50 fade as these operations are removed.
	
	\begin{figure*}[!h]
		\centering
		VGG16 \\
		\includegraphics[width=\textwidth]{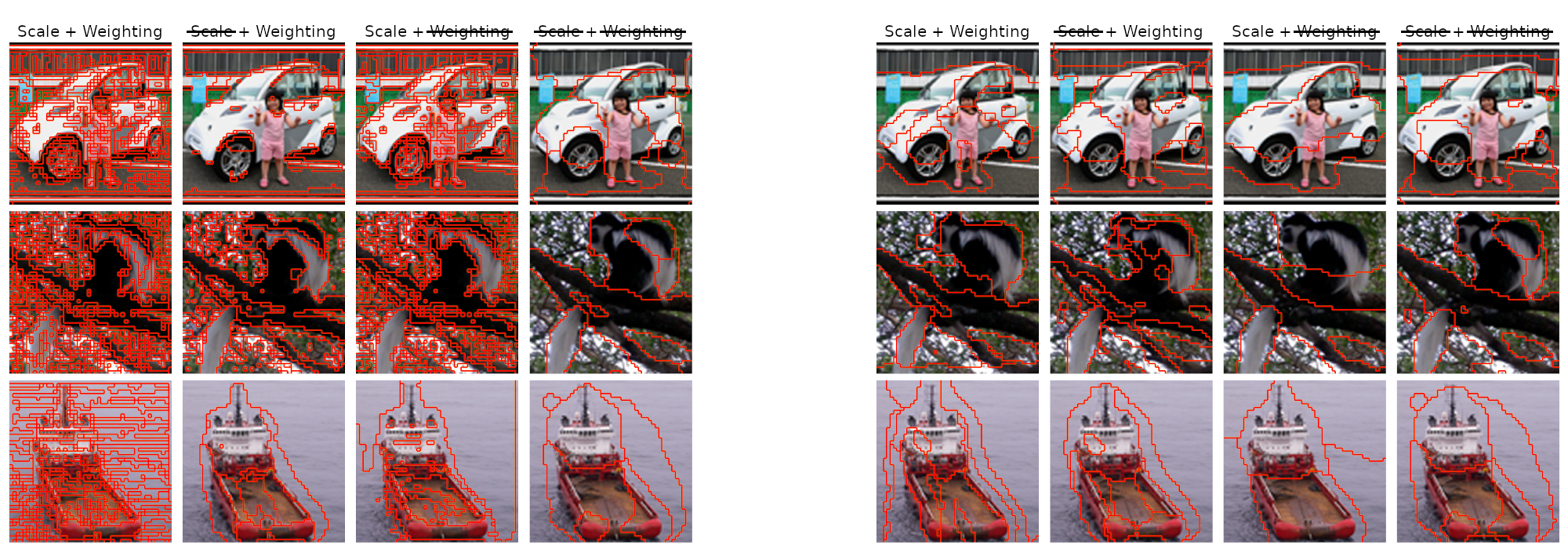}\\ 
		ResNet18 \\
		\includegraphics[width=\textwidth]{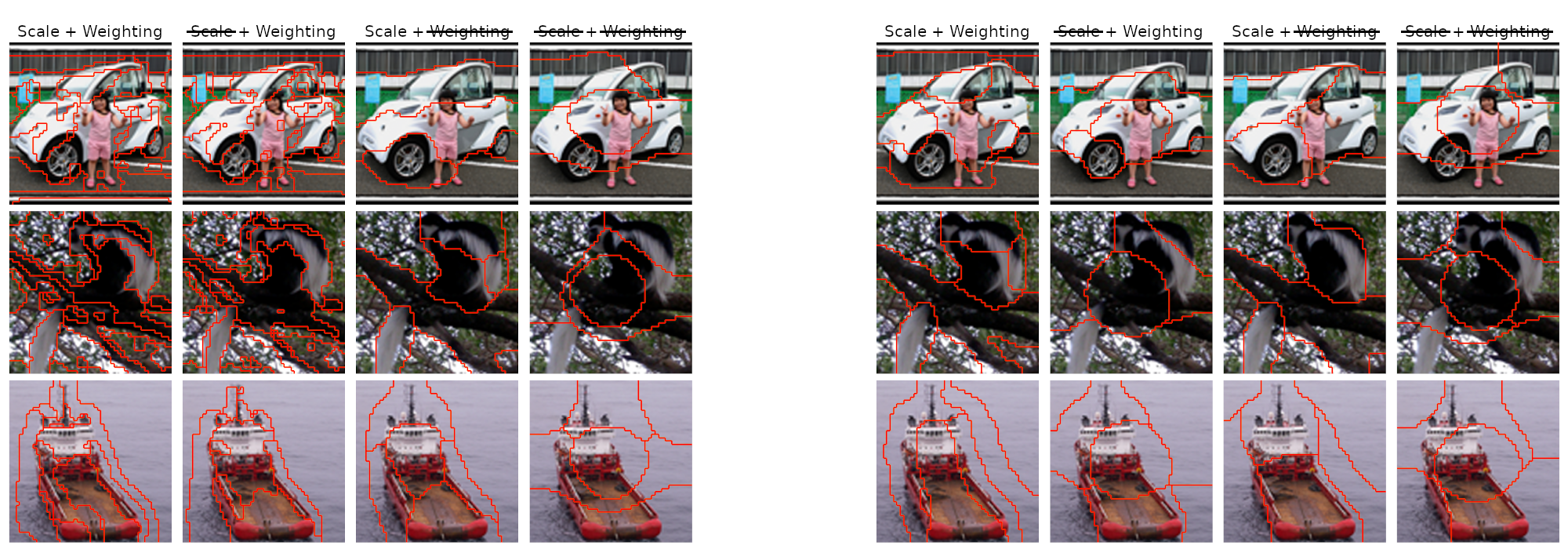}\\ 
		ResNet50 \\
		\includegraphics[width=\textwidth]{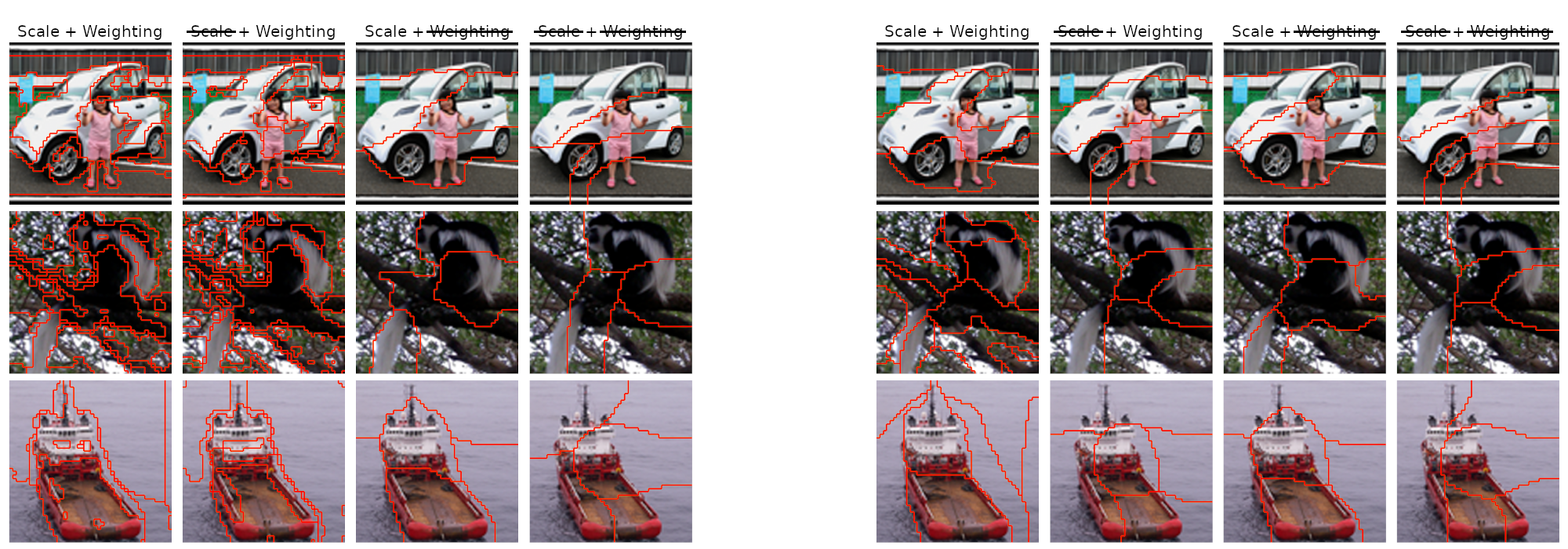}\\ 
		\caption{Ablation study of the scaling and weighting of the feature activations with K=5. On the left, all activations are processed, while on the right, only the last three are used for the segmentation.}
		\label{fig:nrm+wgt}
	\end{figure*}
	
	\section{Consistency of $k$-means}\label{apx-sec:cons}
	
\begin{figure*}[!h]
		\centering
		\includegraphics[width=.7\textwidth]{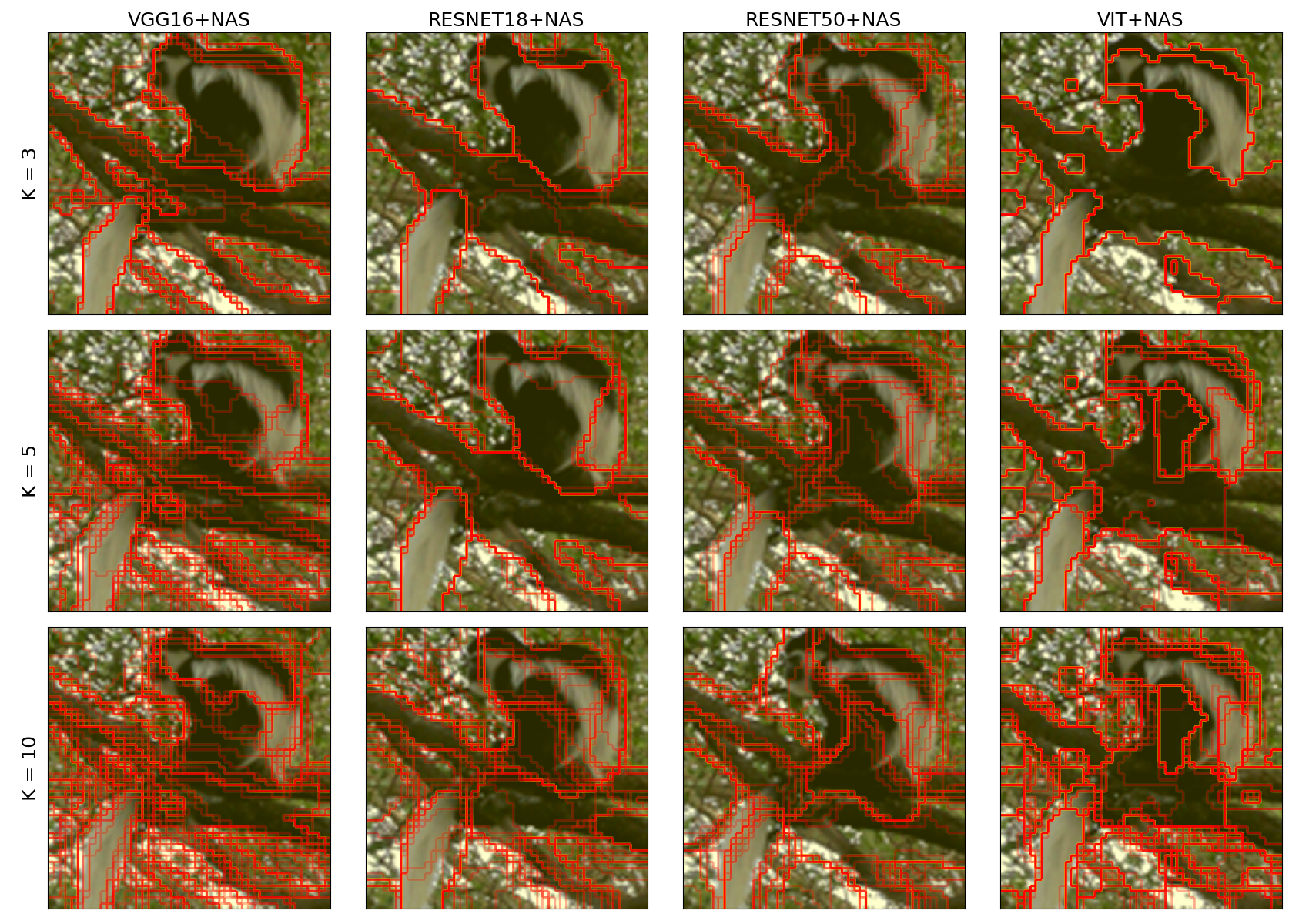}\\
		\caption{The overlay of 10 NAS computed with $k$-means for different architecture and number of clusters reveals a mean segmentation.}
		\label{apx-fig:cons}
	\end{figure*}

	The stochasticity of $k$-means affects the consistency of the superpixels produced by NAS. In Figure~\ref{apx-fig:cons}, we overlay 10 NAS segmentations for different settings. We see that although the border of the superpixels varies, a mean segmentation exists.
	
	\section{Hierarchical Clustering}\label{apx-sec:agg}
	
\begin{figure*}[!h]
		\centering
		\includegraphics[width=.7\textwidth]{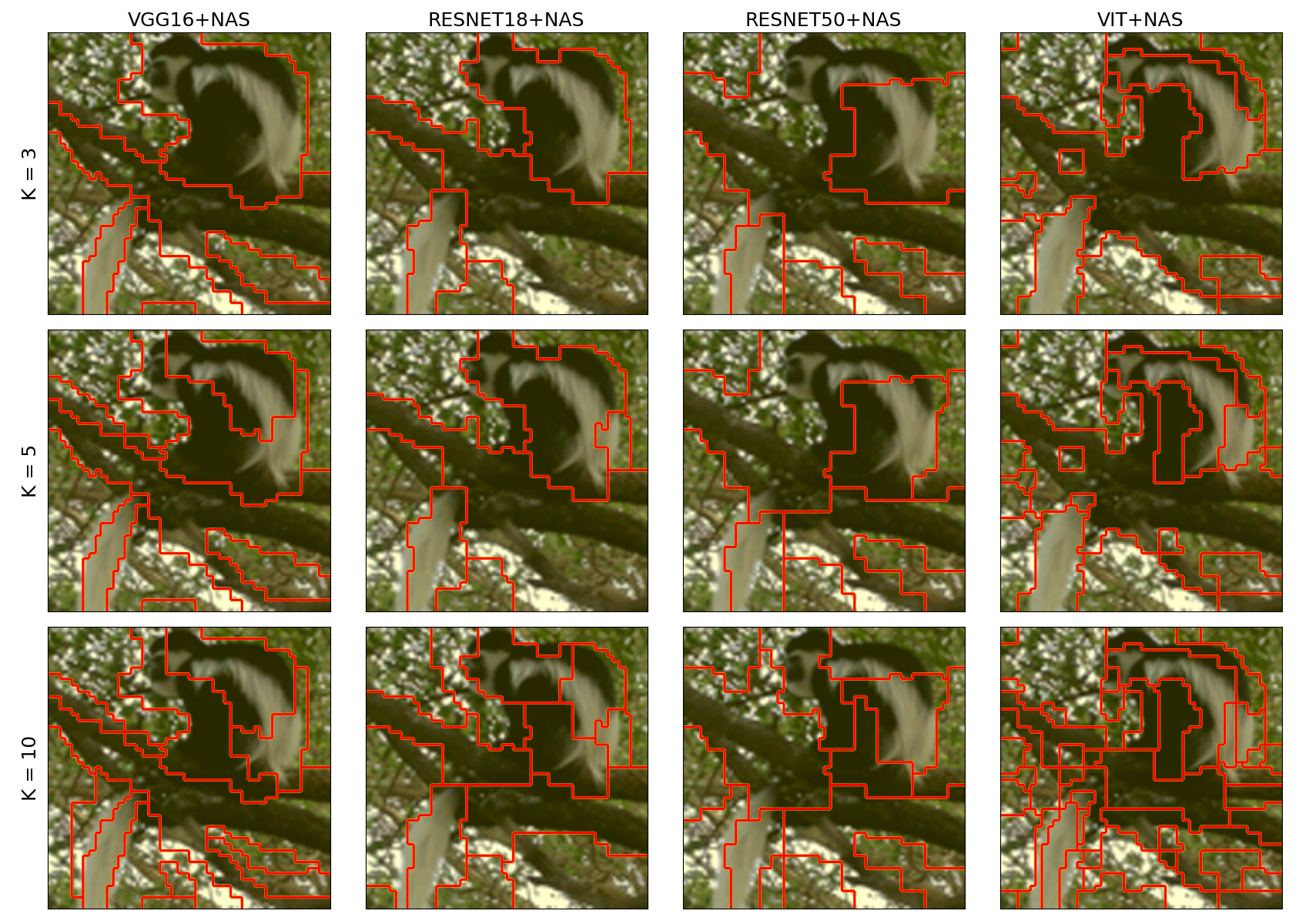}\\
		\caption{Hierarchical clustering with Ward linkage produces consistent superpixels, hence increasing the number of clusters consists of refining the same clustering.}
		\label{apx-fig:agg}
	\end{figure*}

	Hierarchical Clustering is a deterministic method that thus returns consistently the same segmentation. However, it is expensive to compute, hence, we recommend it only for single image analyses. Another strong point for that method is the hierarchy of the clusterings. Namely, we can see in Figure~\ref{apx-fig:agg} that the clusterings within the same column of Figure~\ref{apx-fig:agg} are a refinement of the first one.
	
	\section{ Weakly Supervised Object Localization }\label{apx-sec:wsol}
	We extend in Table~\ref{apx-tab:wsol} with results of Section~\ref{sec:wsol} for VGG16 and ResNet50. 
	The improvement brought by NAS on the IoU@70\% metrics is consistent over the architecture.
	\begin{table*}[!h]
		\vspace{-1em}
		\centering
		\caption{Weakly Supervised Object Detection scores on CUBv2 for VGG16, ResNet18 and ResNet50.}\label{apx-tab:wsol}
		{	\footnotesize \centering
			\begin{tabular*}{\linewidth}{@{\extracolsep{\fill}}p{1.2cm}cccc|cccc|cccc}
				\toprule 
				& \multicolumn{4}{c}{VGG16}& \multicolumn{4}{c}{ResNet18}& \multicolumn{4}{c}{ResNet50}\\
				&  {\tt MaxBox} & IoU 	& IoU 	& IoU\% &  {\tt MaxBox} & IoU 	& IoU 	& IoU\% &  {\tt MaxBox} & IoU 	& IoU 	& IoU\% \\ 
				Methods &  {\tt AccV2}  & @30\% & @50\% & 70\%  &  {\tt AccV2}  & @30\% & @50\% & 70\%  &  {\tt AccV2}  & @30\% & @50\% & 70\% \\ 
				\midrule

				IGrad. 	& 32.0 & 60.2 & 26.1 & 9.7 		& 36.7 & 68.8 & 31.4 & 9.9 		& 38.9 & 72.9 & 33.8 & 9.9 \\
				+SLIC  	& +6.3 & +12.5 & +6.3 & +0.0 	& +1.5  & +5.0 & +0.9  & -1.5	& -0.2 & +2.0 & -1.2 & -1.3 \\
				+NAS   	& +1.3 & +0.7 & +2.2 & +0.9		& +1.5  & -1.1 & +3.3  & +2.4 	& +0.0 & -5.1 & +1.8 & +3.3 \\ \midrule
				
				LRP 	& 71.8 & 93.1 & 77.4 & 44.9 	& 63.6 & 95.4 & 71.3 & 24.0 	& 67.8 & 96.6 & 76.8 & 29.9 \\
				+SLIC 	& -11.0 & +0.3 & -10.9 & -22.4 	& -8.1 & -1.5 & -11.5 & -11.5 	& -11.3 & -2.4 & -15.7 & -15.8 \\
				+NAS 	& -9.5 & -7.2 & -13.2 & -8.2 	& +0.8 & -3.8 & -2.7 & +9.0 	& -4.1 & -6.3 & -9.8 & +3.8 \\ \midrule
				
				GCAM++ 	& 70.6 & 98.8 & 83.9 & 29.2 	& 57.1 & 97.4 & 63.9 & 9.9 		& 57.9 & 98.3 & 64.8 & 10.7 \\
				+SLIC 	& -6.1 & -0.9 & -10.1 & -7.1 	& -1.1 & -1.0 & -3.4 & +1.1 	& -2.0 & -1.6 & -4.6 & +0.2 \\
				+NAS 	& +2.1 & -2.2 & -5.2 & +13.7 	& +8.9 & -3.0 & +8.6 & +21.1 	& +6.4 & -4.2 & +4.8 & +18.5 \\ \midrule
				
				RISE 	& 49.7 & 85.9 & 50.7 & 11.5 	& 32.4 & 65.3 & 26.2 & 5.7 		& 35.4 & 68.0 & 30.8 & 7.4 \\
				+SLIC 	& -0.5 & +2.0  & -2.3 & -1.6 	& +0.7 & +0.9 & +1.2 & +0.1 	& +0.2 & -0.2 & +0.7 & +0.1 \\
				+NAS 	& +6.7 & +1.8 & +10.6  & +26.8 	& +6.4 & +2.6 & +9.4 & +7.0 	& +6.6 & +1.2 & +8.8 & +9.6 \\ \midrule
				
				XRAI 	& 54.4 & 90.8 & 60.4 & 18.1 	& 31.3 & 63.2 & 24.9 & 6.0 		& 29.8 & 60.7 & 23.2 & 5.5 \\
				+SLIC 	& +0.6  & +1.0  & -2.5 & -2.8	& +0.9 & +0.8 & +1.6 & +0.2 	& +0.7 & +0.5 & +1.7 & -0.1 \\
				+NAS 	& +9.1  & -0.2 &  +6.6 & +14.8 	& +4.9 & +2.0 & +7.4 & +5.3 	& +5.0 & +1.3 & +7.9 & +5.8 \\ 
				
				\bottomrule

		\end{tabular*}}
	\end{table*}

	\section{Runtime}
	
	We report in Table~\ref{apx-tab:run} runtimes for our algorithm and for SLIC and Felzenszwalb. These two rely on the skimage implementation.
	We evaluate for different architectures, depth of the feature activation, and number of clusters.
	For the sake of completeness, we also compute NAS with a hierarchical clustering with Ward linkage on the combination: ResNet18, depth [2,3,4] and K=5. The time is given between brackets in the corresponding cell of the table.
	We recall that we ran this experiment on an Nvidia L40S 45GB supported with an AMD 7452 32-core CPU with 500G of RAM. SLIC and Felzenszwalb use only the CPU.
	We report averages computed over 1000 train images from STL10.

	\begin{table*}[!h]
		\vspace{-1em}
		\centering
		\caption{Runtime for NAS, SLIC, and Felzenszwalb segmentation. Times are given in ms. Statistically fastest runtimes are marked in bold. The fastest runtime for NAS is underlined.}\label{apx-tab:run}
		{	\footnotesize \centering
			\begin{tabular*}{\linewidth}{@{\extracolsep{\fill}}cccccccc}
				\toprule 
				
				& & VGG16 & {ResNet18} & {ResNet50}\\ \midrule
				\multirow{2}*{[0,1,2,3,4]}  & K = 5  & $16.6$ & $15.7$  & $17.4$ & & \multirow{2}*{SLIC}  & \multirow{2}*{{$2.6$}} \\
				& K = 10 & $19.6$ & $19.9$ & $27.2$ \\ \midrule
				\multirow{2}*{[2,3,4]}  & K = 5  & \underline{$11.9$} & $13.9$ ($833.3$) & $14.9$ & & \multirow{2}*{Felzenszwalb}  & \multirow{2}*{{ $\bf 2.3$}} \\
				& K = 10 & $16.4$ & $17.9$ & $19.9$ \\
				\bottomrule	
		\end{tabular*}}
	\end{table*}
	
	Both SLIC and Felzenszwalb are faster by an order of magnitude than our method. The different combinations of architecture, the depth of the activation, and the number of clusters return runtimes of the same order. The combination VGG16, [2,3,4], and K=5 is the fastest but also the one with the fewest parameters.
	Finally, NAS based on hierarchical clustering is $60$ times slower than if combined with $k$-means.

\end{document}